\def\year{2020}\relax
\documentclass[letterpaper]{article} 
\usepackage{aaai20}  
\usepackage{times}  
\usepackage{helvet} 
\usepackage{courier}  
\usepackage[hyphens]{url}  
\usepackage{multirow}
\usepackage{graphicx} 
\usepackage{graphicx}  
\usepackage[utf8]{inputenc} 
\usepackage[T1]{fontenc}    
\usepackage[draft]{hyperref}       
\usepackage{url}            
\usepackage{booktabs}       
\usepackage{amsfonts}       
\usepackage{nicefrac}       
\usepackage{microtype}      
\usepackage{amssymb,amsmath,amsthm}
\usepackage{tikz}
\usepackage{diagbox}
\usepackage{cleveref}
\usepackage{subfigure}

\usepackage[linesnumbered,ruled,vlined]{algorithm2e}
\usepackage{xcolor}
\newcommand{\citet}[1]{\citeauthor{#1}~\shortcite{#1}}
\newcommand{\citep}{\cite}

\title{Variational Metric Scaling for Metric-Based Meta-Learning}

\author{Jiaxin Chen,
	Li-Ming Zhan,
	Xiao-Ming Wu\thanks{Corresponding authors.},
	Fu-lai Chung$^\ast$ \\
	Department of Computing\\ The Hong Kong Polytechnic University\\
	\{jiax.chen, lmzhan.zhan\}@connect.polyu.hk,
	xiao-ming.wu@polyu.edu.hk,
	cskchung@comp.polyu.edu.hk }

\begin{document}
	\maketitle
		\begin{abstract}
		Metric-based meta-learning has attracted a lot of attention due to its effectiveness and efficiency in few-shot learning. Recent studies show that metric scaling plays a crucial role in the performance of metric-based meta-learning algorithms. However, there still lacks a principled method for learning the metric scaling parameter automatically. In this paper, we recast metric-based meta-learning from a Bayesian perspective and develop a variational metric scaling framework for learning a proper metric scaling parameter. Firstly, we propose a stochastic variational method to learn a single global scaling parameter. To better fit the embedding space to a given data distribution, we extend our method to learn a dimensional scaling vector to transform the embedding space. Furthermore, to learn task-specific embeddings, we generate task-dependent dimensional scaling vectors with amortized variational inference. Our method is end-to-end without any pre-training and can be used as a simple plug-and-play module for existing metric-based meta-algorithms. Experiments on \textit{mini}ImageNet show that our methods can be used to consistently improve the performance of existing metric-based meta-algorithms including prototypical networks and TADAM. The source code can be downloaded from \textcolor{blue}{https://github.com/jiaxinchen666/variational-scaling}.
 
	\end{abstract}

	\section{1. Introduction}
	Few-shot learning \cite{li2006one} aims to assign unseen samples (\textit{query}) to the belonging categories with very few labeled samples (\textit{support}) in each category. 
	A promising paradigm for few-shot learning is meta-learning, which learns general patterns from a large number of tasks for fast adaptation to unseen tasks. Recently, metric-based meta-learning algorithms \cite{garcia2017few,koch2015siamese,snell2017prototypical,vinyals2016matching} demonstrate great potential in few-shot classification. Typically, they learn a general mapping, which projects queries and supports into an embedding space.
	These models are trained in an episodic manner \cite{vinyals2016matching} by minimizing the distances between a query and same-labeled supports in the embedding space. Given a new task in testing phase, a nearest neighbour classifier is applied to assign a query to its nearest class in the embedding space.
	
Many metric-based meta-algorithms (short form for meta-learning algorithms) employ a softmax classifier with cross-entropy loss, which is computed with the logits being the distances between a query and supports in the embedding (metric) space. However, it has been shown that the scale of the logits -- the metric scaling parameter, is critical to the performance of the learned model. \citet{snell2017prototypical} found that Euclidean distance significantly outperforms cosine similarity in few-shot classification, while \citet{oreshkin2018tadam} and \citet{wang2018large} pointed out that there is no clear difference between them if the logits are scaled properly. They supposed that there exists an optimal metric scaling parameter which is data and architecture related, but they only used cross validation to manually set the parameter, which requires pre-training and cannot find an ideal solution. 

In this paper, we aim to design an end-to-end method that can automatically learn an accurate metric scaling parameter. Given a set of training tasks, to learn a data-dependent metric scaling parameter that can generalize well to a new task, Bayesian posterior inference over learnable parameters is a theoretically attractive framework \cite{gordon2018meta,ravi2018amortized}. We propose to recast metric-based meta-algorithms from a Bayesian perspective and take the metric scaling parameter as a global parameter. As exact posterior inference is intractable, we introduce a variational approach to efficiently approximate the posterior distribution with stochastic variational inference.

    While a proper metric scaling parameter can improve classification accuracy via adjusting the cross-entropy loss, it simply rescales the embedding space but does not change the relative locations of the embedded samples. To transform the embedding space to better fit the data distribution, we propose a dimensional variational scaling method to learn a scaling parameter for each dimension, i.e., a metric scaling vector. Further, in order to learn task-dependent embeddings \cite{oreshkin2018tadam}, we propose an amortized variational approach to generate task-dependent metric scaling vectors, accompanied by an auxiliary training strategy to avoid time-consuming pre-training or co-training.
    
    Our metric scaling methods can be used as pluggable modules for metric-based meta-algorithms. For example, it can be incorporated into prototypical networks (PN) \cite{snell2017prototypical} and all PN-based algorithms to improve their performance. To verify this, we conduct extensive experiments on the \textit{mini}ImageNet benchmark for few-shot classification progressively. First, we show that the proposed stochastic variational approach consistently improves on PN, and the improvement is large for PN with cosine similarity. Second, we show that the dimensional variational scaling method further improves upon the one with single scaling parameter, and the task-dependent metric scaling method with amortized variational inference achieves the best performance. We also incorporate the dimensional metric scaling method into TADAM \cite{oreshkin2018tadam} in conjunction with other tricks proposed by the authors and observe notable improvement. Remarkably, after incorporating our method, TADAM achieves highly competitive performance compared with state-of-the-art methods. 
        
   To sum up, our contributions are as follows:
   \begin{itemize}
   	\item We propose a generic variational approach to automatically learn a proper metric scaling parameter for metric-based meta-algorithms. 
   	\item We extend the proposed approach to learn dimensional and task-dependent metric scaling vectors to find a better embedding space by fitting the dataset at hand.
   	\item As a pluggable module, our method can be efficiently used to improve existing metric-based meta-algorithms.
   \end{itemize}
    
	\section{2. Related Work}
\subsubsection{Metric-based meta-learning.} 
	\citet{koch2015siamese} proposed the first metric-based meta-algorithm for few-shot learning, in which a siamese network~\cite{chopra2005learning} is trained with the triplet loss to compare the similarity between a query and supports in the embedding space. Matching networks \cite{vinyals2016matching} proposed the episodic training strategy and used the cross-entropy loss where the logits are the distances between a query and supports. Prototypical networks~\cite{snell2017prototypical} improved Matching networks by computing the distances between a query and the prototype (mean of supports) of each class. Many metric-based meta-algorithms~\cite{oreshkin2018tadam,fort2017gaussian,sung2018learning,li2019finding} extended prototypical networks in different ways.
	
	Some recent methods proposed to improve prototypical networks by extracting task-conditioning features. \citet{oreshkin2018tadam} trained a network to generate task-conditioning parameters for batch normalization. \citet{li2019finding} extracted task-relevant features with a category traversal module. Our methods can be incorporated into these methods to improve their performance.
		
In addition, there are some works related to our proposed dimensional scaling methods. \citet{kang2018few}
trained a meta-model to re-weight features obtained from the base feature extractor and applied it for few-shot object detection. 
\citet{lai2018task} proposed a generator to generate task-adaptive weights to re-weight the embeddings, which can be seen as a special case of our amortized variational scaling method.

\subsubsection{Metric scaling.} 
Cross-entropy loss is widely used in many machine learning problems, including metric-based meta-learning and metric learning
\cite{babenko2015aggregating,ranjan2017l2,liu2017sphereface,wang2017normface,zhang2018heated,babenko2015aggregating,wan2018rethinking}. In metric learning, the influence of metric scaling on the cross-entropy loss was first studied in \citet{wang2017normface} and \citet{ranjan2017l2}. They treated the metric scaling parameter as a trainable parameter updated with model parameters or a fixed hyperparameter. \citet{zhang2018heated} proposed a ``heating-up'' scaling strategy, where the metric scaling parameter decays manually during the training process.  
The scaling of logits in cross-entropy loss for model compression was also studied in \citet{hinton2015distilling}, where it is called temperature scaling. The temperature scaling parameter has also been used in confidence  calibration~\cite{guo2017calibration}.
	
The effect of metric scaling for few-shot learning was first discussed in \citet{snell2017prototypical} and \citet{oreshkin2018tadam}. The former found that Euclidean distance outperforms cosine similarity significantly in prototypical networks, and the latter argued that the superiority of Euclidean distance could be offset by imposing a proper metric scaling parameter on cosine similarity and using cross validation to select the parameter.

	\section{3. Preliminaries}
	    \subsection{3.1. Notations and Problem Statement}
	Let $\mathcal{Z}=\mathcal{X}\times\mathcal{Y}$ be a domain where $\mathcal{X}$ is the input space and $\mathcal{Y}$ is the output space. Assume we observe a meta-sample $\mathbf{S}=\{\mathcal{D}_i=\mathcal{D}_i^{tr}\cup\mathcal{D}_i^{ts}\}_{i=1}^n$ including $n$ training tasks, where the $i$-{th} task consists of a support set of size $m$, $\mathcal{D}_i^{tr}=\{z_{i,j}=(x_{i,j},y_{i,j})\}_{j=1}^m$, and a query set of size $q$, $\mathcal{D}_i^{ts}=\{z_{i,j}=(x_{i,j},y_{i,j})\}_{j=m+1}^{m+q}$. Each training data point $z_{i,j}$ belongs to the domain $\mathcal{Z}$. Denote by $\theta$ the model parameters and $\alpha$ the metric scaling parameter. Given a new task and a support set $\mathcal{D}^{tr}$ sampled from the task, the goal is to predict the label $y$ of a query $x$.
	\subsection{3.2. Prototypical Networks}
	Prototypical networks (PN) \cite{snell2017prototypical} is a popular and highly effective metric-based meta-algorithm. PN learns a mapping $\phi_\theta$ which projects queries and the supports to an $M$-dimensional embedding space. For each class $k\in\{1,2,\dots,K\}$, the mean vector of the supports of class $k$ in the embedding space is computed as the class \textit{prototype} $\mathbf{c}_k$. The embedded query is compared with the prototypes and assigned to the class of the nearest prototype. Given a similarity metric $d:\mathbb{R}^M\times \mathbb{R}^M\rightarrow \mathbb{R}^{+}$, the probability of a query $z_{i,j}$ belonging to class $k$ is,
	\begin{align}
			p_\theta(y_{i,j}=k|x_{i,j},\mathcal{D}_i^{tr})=\frac{e^{-d(\phi_\theta(x_{i,j}),\mathbf{c}_k)}}{\sum_{{k}'=1}^{K}e^{-d(\phi_\theta(x_{i,j}),\mathbf{c}_{{k}'})}}.
	\end{align}
    Training proceeds by minimizing the cross-entropy loss, i.e., the negative log-probability $-\log p_\theta(y_{i,j}=k|x_{i,j},\mathcal{D}_i^{tr})$ of its true class $k$. After introducing the metric scaling parameter $\alpha$, the classification loss of the $i^{th}$ task becomes
    \begin{align}
    	\mathcal{L}(\theta;\mathcal{D}_i)=-\sum_{j=m+1}^{m+q}\log\frac{e^{-\alpha* d(\phi_\theta(x_{i,j}),\mathbf{c}_{y_{i,j}})}}{\sum_{{k}'=1}^{K}e^{-\alpha* d(\phi_\theta(x_{i,j}),\mathbf{c}_{{k}'})}}.
    \end{align}
	The metric scaling parameter $\alpha$ has been found to affect the performance of PN significantly.
	\section{4. Variational Metric Scaling}
	\subsection{4.1. Stochastic Variational Scaling}~\label{stochastic}
	In the following, we recast metric-based meta-learning from a Bayesian perspective. 
	The predictive distribution can be parameterized as 
	\begin{align}~\label{predictive_original}
	p_\theta(y|x,\mathcal{D}^{tr},\mathbf{S})=\int p_\theta(y|x,\mathcal{D}^{tr},\alpha)p_\theta(\alpha|\mathbf{S})d\alpha.
	\end{align}
	The conditional distribution $p_\theta(y|x,\mathcal{D}^{tr},\alpha)$ is the discriminative classifier parameterized by $\theta$. Since the posterior distribution $p_\theta(\alpha|\mathbf{S})$ is intractable, we propose a variational distribution $q_\psi(\alpha)$ parameterized by parameters $\psi$ to approximate $p_\theta(\alpha|\mathbf{S})$. 
	By minimizing the KL divergence between the approximator $q_\psi(\alpha)$ and the real posterior distribution $p_\theta(\alpha|\mathbf{S})$, we obtain the objective function
	\begin{align}~\label{KL}
		&\mathcal{L}(\psi,\theta;\mathbf{S})=\int q_{\psi}(\alpha)\log\frac{q_{\psi}(\alpha)}{p_{\theta}(\alpha|\mathbf{S})}d\alpha\nonumber\\
		=&-\int q_{\psi}(\alpha)\log\frac{p_\theta(\mathbf{S}|\alpha)p(\alpha)}{q_{\psi}(\alpha)}d\alpha+\log p(\mathbf{S})\nonumber\\
		=&-\int q_{\psi}(\alpha)\log p_\theta(\mathbf{S}|\alpha)d\alpha+KL(q_{\psi}(\alpha)|p(\alpha))+\text{const}\nonumber\\
		=&-\sum_{i=1}^{n}\sum_{j=m+1}^{m+q}\int q_{\psi}(\alpha)\log p_\theta(y_{i,j}|x_{i,j},\mathcal{D}_i^{tr},\alpha)d\alpha\nonumber\\
		&+KL(q_{\psi}(\alpha)|p(\alpha))+\text{const}.
	\end{align}	
	We want to optimize $\mathcal{L}(\psi,\theta;\mathbf{S})$ w.r.t. both the model parameters $\theta$ and the variational parameters $\psi$. The gradient and the optimization procedure of the model parameters $\theta$ are similar to the original metric-based meta-algorithms~\cite{vinyals2016matching,snell2017prototypical} as shown in Algorithm~\ref{Algorithm1}. 
	
	To derive the gradients of the variational parameters, we leverage the re-parameterization trick proposed by \citet{kingma2013auto} to derive a practical estimator of the variational lower bound and its derivatives w.r.t. the variational parameters. In this paper, we use this trick to estimate the derivatives of $\mathcal{L}(\psi,\theta;\mathbf{S})$ w.r.t. $\psi$. For a distribution $q_{\psi}(\alpha)$, we can re-parameterize $\alpha\sim q_{\psi}(\alpha)$ using a differentiable transformation $\alpha=g_{\psi}(\epsilon)$, if exists, of an auxiliary random variable $\epsilon$. For example, given a Gaussian distribution $q_{\mu,\sigma}(\alpha)=\mathcal{N}(\mu,\sigma^2)$, the re-parameterization is $g_{\mu,\sigma}(\epsilon)=\epsilon\sigma+\mu$, where $\epsilon\sim\mathcal{N}(0,1)$. 
	Hence, the first term in (\ref{KL}) is formulated as 
	$-\sum_{i=1}^{n}\sum_{j=m+1}^{m+q}\mathbb{E}_{\epsilon\sim p(\epsilon)}\log p_\theta(y_{i,j}|x_{i,j},\mathcal{D}_i^{tr},g_\psi(\epsilon))$. 
	
	We apply a Monte Carlo integration with a single sample $\alpha_i=g_\psi(\epsilon_i)$ for each task to get an unbiased estimator. Note that $\alpha_i$ is sampled for the task $\mathcal{D}_i$ rather than for each instance, i.e., $\{z_{i,j}\}_{j=1}^{m+q}$ share the same $\alpha_i$.
	The second term in (\ref{KL}) can be computed with a given prior distribution $p(\alpha)$. Then, the final objective function is
	\begin{align}~\label{finalloss}
	\mathcal{L}(\psi,\theta;\mathbf{S})&=-\sum_{i=1}^{n}\sum_{j=m+1}^{m+q}\log p_\theta(y_{i,j}|x_{i,j},\mathcal{D}_i^{tr},g_\psi(\epsilon_i))\nonumber\\
	&+KL(q_{\psi}(\alpha)|p(\alpha))
	\end{align}
	\subsubsection{ Estimation of gradients.}
	The objective function~(\ref{finalloss}) is a general form. Here, we consider $q_{\psi}(\alpha)$ as a Gaussian distribution
	$q_{\mu,\sigma}(\alpha)=\mathcal{N}(\mu,\sigma^2)$. The prior distribution is also a Gaussian distribution $p(\alpha)=\mathcal{N}(\mu_0,\sigma_0^2)$. By the fact that the KL divergence of two Gaussian distributions has a closed-form solution, we obtain the following objective function \begin{align}
	\mathcal{L}(\mu,\sigma,\theta;\mathbf{S})&=-\sum_{i=1}^{n}\sum_{j=m+1}^{m+q}\log p_\theta(y_{i,j}|x_{i,j},\mathcal{D}_i^{tr},g_{\mu,\sigma}(\epsilon_i))\nonumber\\
	&+\log\frac{\sigma_0}{\sigma}+\frac{\sigma^2+(\mu-\mu_0)^2}{2\sigma_0^2},
	\end{align}
	where $g_{\mu,\sigma}(\epsilon_i)=\sigma\epsilon_i+\mu$.
	The derivatives of $\mathcal{L}(\mu,\sigma,\theta;\mathbf{S})$ w.r.t. $\mu$ and $\sigma$ respectively are 
	\begin{align}
	&\frac{\partial\mathcal{L}(\mu,\sigma,\theta;\mathbf{S})}{\partial\mu}=\nonumber\\
	&-\sum_{i=1}^{n}\sum_{j=m+1}^{m+q}\frac{\partial\log p_\theta(y_{i,j}|x_{i,j},\mathcal{D}_i^{tr},g_{\mu,\sigma}(\epsilon_i))}{\partial g_{\mu,\sigma}(\epsilon_i)}+\frac{\mu-\mu_0}{\sigma_0^2}~\label{gradient of mu},\\
	&\frac{\partial\mathcal{L}(\mu,\sigma,\theta;\mathbf{S})}{\partial\sigma}=\nonumber\\
	&-\sum_{i=1}^{n}\sum_{j=m+1}^{m+q}\frac{\partial\log p_\theta(y_{i,j}|x_{i,j},\mathcal{D}_i^{tr},g_{\mu,\sigma}(\epsilon_i))}{\partial g_{\mu,\sigma}(\epsilon_i)}*\epsilon_i\nonumber\\
	&-\frac{1}{\sigma}+\frac{\sigma}{\sigma_0^2}~\label{gradient of sigma}. 
	\end{align}
	In particular, we apply the proposed variational metric scaling method to Prototypical Networks with feature extractor $\phi_\theta$. The details of the gradients and the iterative update procedure are shown in Algorithm~\ref{Algorithm1}. It can be seen that the gradients of the variational parameters are computed using the intermediate quantities in the computational graph of the model parameters $\theta$ during back-propagation, hence the computational cost is very low.

    For meta-testing, we use $\mu$ (mean) as the metric scaling parameter for inference. 

\SetKwInput{KwInput}{Input}               
\SetKwInput{KwOutput}{Output}
\SetKwInput{KwInitialize}{Initialize}             
	
\begin{algorithm}[h]
	\DontPrintSemicolon		
	\KwInput{Meta-sample $\{\mathcal{D}_i\}_{i=1}^n$, learning rates $l_\theta$, $l_\psi$ and $\mu_0,\sigma_0$.}
	 \KwInitialize{$\mu,\sigma$ and $\theta$ randomly.}
	\For{$i$ in $\{1,2,\dots,n\}$}
		{\textcolor{blue}{$\epsilon_i\sim\mathcal{N}(0,1),\alpha_i=\sigma\epsilon_i+\mu$} 
			
		\tcp*{Sample $\alpha_i$ for $i^{th}$ task.}
		\For{$k$ in $\{1,2,\dots,K\}$}
			{	
			$\textbf{c}_k=\frac{1}{N}\sum_{z_{i,j}\in\mathcal{D}_i^{tr},y_{i,j}=k}\phi_\theta(x_{i,j})$ 
			
			\tcp*{Compute prototypes.}}
		\For{$j$ in $\{m+1,2,\dots,m+q\}$}
			{$d(x_{i,j},\mathbf{c}_{k})=\| \phi_\theta(x_{i,j})-\mathbf{c}_k \|^2_2$
				
			$p(y_{i,j}=k)=\frac{e^{-\textcolor{blue}{\alpha_i}*d(x_{i,j},\mathbf{c}_k)}}{\sum_{{k}'=1}^{K}e^{-\textcolor{blue}{\alpha_i}*d(x_{i,j},\mathbf{c}_{{k}'})}}$}
			
			$\theta=\theta-l_\theta*\nabla_\theta \mathcal{L}(\mu,\sigma,\theta;\mathcal{D}_i)$

			\tcp*{Update the model parameters $\theta$.}
			
			\textcolor{blue}{$\mu=\mu-l_\psi*(\sum_{j=m+1}^{m+q}(-d(x_{i,j},\mathbf{c}_{y_{i,j}})+\sum_{{k}'=1}^{K}p(y_{i,j}={k}')*d(x_{i,j},\mathbf{c}_{{k}'}))+\frac{\mu-\mu_0}{\sigma_0^2})$}
			
			\textcolor{blue}{$\sigma=\sigma-l_\psi*(\sum_{j=m+1}^{m+q}\epsilon_i*(-d(x_{i,j},\mathbf{c}_{y_{i,j}})+\sum_{{k}'=1}^{K}p(y_{i,j}={k}')*d(x_{i,j},\mathbf{c}_{{k}'}))-\frac{1}{\sigma}+\frac{\sigma}{\sigma_0^2})$}
			
			\tcp*{Update the variational parameters $\psi=\{\mu,\sigma\}$.}}
		\caption{Stochastic Variational Scaling for Prototypical Networks}
		\label{Algorithm1}
	\end{algorithm}

\begin{figure*}
	\centering
	\begin{tikzpicture}
	\pgfmathparse{sqrt(2)}
	\draw[step=0.5,color=gray!20] (-1,-1) grid (1,1); 
	\draw[step=0.5,color=gray!20] (3,-1) grid (5,1);  
	\draw[step=0.5,color=gray!20] (8.5,-1) grid (10.5,1);  
	\draw[->] (-1,0) -- (1,0); 
	\draw[->] (0,-1) -- (0,1); 
	\draw (0,0) circle (0.75); 
	\node[circle,fill=red,inner sep=0pt,minimum size=4pt,label={[xshift=0.2cm, yshift=-0.1cm]$\mathbf{c}_1$}] (c1) at (0.75*0.5*\pgfmathresult,0.75*0.5*\pgfmathresult) {};
    \node[circle,fill=blue,inner sep=0pt,minimum size=4pt,label=right:{$Q$}] (q1) at ((0.75*0.5*\pgfmathresult,-0.75*0.5*\pgfmathresult) {};
    \node[circle,fill=red,inner sep=0pt,minimum size=4pt,label={[xshift=-0.2cm, yshift=-0.05cm]$\mathbf{c}_2$}] (c2) at (0,-0.75) {};
    \draw[dash pattern=on 2pt off 1pt](c1)--(q1);
    \draw[dash pattern=on 2pt off 1pt,red](c2)--(q1);
	\draw[step=0.5,color=gray!40] (3,-1) grid (5,1);  
	\draw[->] (3,0) -- (5,0); 
	\draw[->] (4,-1) -- (4,1); 
	\draw (4,0) circle (0.5);  
	\node[circle,fill=red,inner sep=0pt,minimum size=4pt,label={[xshift=0.2cm, yshift=-0.1cm]$\mathbf{c}_1$}] (c21) at (4+0.25*\pgfmathresult,0.25*\pgfmathresult) {};
	\node[circle,fill=blue,inner sep=0pt,minimum size=4pt,label=right:{$Q$}] (q2) at (4+0.25*\pgfmathresult,-0.25*\pgfmathresult) {};
	\node[circle,fill=red,inner sep=0pt,minimum size=4pt,label={[xshift=-0.2cm, yshift=-0.5cm]$\mathbf{c}_2$}] (c22) at (4,-0.5) {};
	\draw[dash pattern=on 2pt off 1pt](c21)--(q2);
	\draw[dash pattern=on 2pt off 1pt,red](c22)--(q2);
	\draw[->] (8.5,0) -- (10.5,0); 
	\draw[->] (9.5,-1) -- (9.5,1); 
	\draw (9.5,0) ellipse (0.75 and 0.25);   
	\draw [<-](1.25,0) -- node [pos=0.5,above,sloped]{$\alpha=1.5$}(2.75,0);
	\draw [->](5.25,0) -- node [pos=0.5,above,sloped]{$(\alpha^1,\alpha^2)=(1.5,0.5)$}(8.25,0);
	\node[circle,fill=red,inner sep=0pt,minimum size=4pt,label={[xshift=0.2cm, yshift=-0.1cm]$\mathbf{c}_1$}] (c31) at (9.5+0.75*0.5*\pgfmathresult,0.25*0.5*\pgfmathresult) {};
	\node[circle,fill=blue,inner sep=0pt,minimum size=4pt,label={[xshift=0.2cm, yshift=-0.6cm]{$Q$}}] (q3) at (9.5+0.75*0.5*\pgfmathresult,-0.25*0.5*\pgfmathresult) {};
	\node[circle,fill=red,inner sep=0pt,minimum size=4pt,label={[xshift=-0.2cm, yshift=-0.5cm]$\mathbf{c}_2$}] (c32) at (9.5,-0.25) {};
	\draw[dash pattern=on 2pt off 1pt](c32)--(q3);
	\draw[dash pattern=on 1pt off 0.5pt,red](c31)--(q3);
	\end{tikzpicture}
	\caption{The middle figure shows a metric space in which the query (blue) and the support samples (red) are normalized to a unit ball. The left and right figures show the spaces scaled by a single parameter $\alpha=1.5$ and a two-dimensional vector $(\alpha^1,\alpha^2)=(1.5,0.5)$, respectively. The query $Q$ is still assigned to class $2$ in the left figure but to class $1$ in the right one.}
	\label{dimension_figure}
\end{figure*}
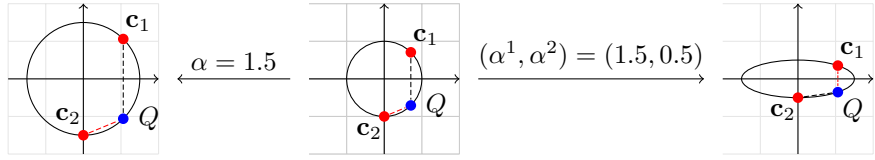
	The proposed variational scaling framework is general. Note that training the scaling parameter $\alpha$ together with the model parameters  \cite{ranjan2017l2} is a special case of our framework, when $q_{\psi}(\alpha)$ is defined as $\mathcal{N}(\mu,0)$, the variance of the prior distribution is $\sigma_0\rightarrow\infty$, and the learning rate is fixed as $l_\theta=l_\psi$.

	\subsection{4.2. Dimensional Stochastic Variational Scaling}~\label{dimension_stochastic}	
	Metric scaling can be seen as a transformation of the metric (embedding) space. Multiplying the distances with the scaling parameter accounts to re-scaling the embedding space. By this point of view, we generalize the single scaling parameter to a dimensional scaling vector which transforms the embedding space to fit the data. 
	
	If the dimension of the embedding space is too low, the data points cannot be projected to a linearly-separable space. Conversely, if the dimension is too high, there may be many redundant dimensions. The optimal number of dimensions is data-dependent and difficult to be selected as a hyperparameter before training.	Here, we address this problem by learning a data-dependent dimensional scaling vector to modify the embedding space, i.e., learning different weights for each dimension to highlight the important dimensions and reduce the influence of the redundant ones. Figure \ref{dimension_figure} shows a two-dimensional example. It can be seen that the single scaling parameter $\alpha$ simply changes the scale of the embedding space, but the dimensional scaling $\alpha=(\alpha^1,\alpha^2)$ changes the relative locations of the query and the supports.  

    The proposed dimensional stochastic variational scaling method is similar to Algorithm~\ref{Algorithm1}, with the variational parameters $\mu=(\mu^1,\mu^2,\dots,\mu^M)$ and $\sigma=(\sigma^1,\sigma^2,\dots,\sigma^M)$. Accordingly, the metric scaling operation is changed to
    \begin{align}
    	&d(x_{i,j},\mathbf{c}_k)\nonumber\\
    	=&(\phi_\theta(x_{i,j})-\mathbf{c}_{k})^{T}\bigl(\begin{smallmatrix}
    	\alpha_i^1 &  &  & \\ 
    	&  \alpha_i^2&  & \\ 
    	&  & \dots & \\ 
    	&  &  & \alpha_i^M
    	\end{smallmatrix}\bigr)(\phi_\theta(x_{i,j})-\mathbf{c}_{k}).
    \end{align}
    The gradients of the variational parameters are still easy to compute and the computational cost can be ignored. 
    
 \subsection{4.3. Amortized Variational Scaling}
The proposed stochastic variational scaling methods above consider the metric scale as a global scalar or vector parameter, i.e., the entire meta-sample $\mathbf{S}=\{\mathcal{D}_i\}_{i=1}^n$ shares the same embedding space. However, the tasks randomly sampled from the task distribution may have specific task-relevant feature representations \cite{kang2018few,lai2018task,li2019finding}. To adapt the learned embeddings to the task-specific representations, we propose to apply amortized variational inference to learn the task-dependent dimensional scaling parameters. 

For amortized variational inference, $\alpha$ is a local latent variable dependent on $\mathcal{D}$ instead of a global parameter.
Similar to stochastic variational scaling, we apply the variational distribution $q_{\psi(\beta)}(\alpha|\mathcal{D})$ to approximate the posterior distribution $p_\theta(\alpha|\mathcal{D})$. In order to learn the dependence between $\alpha$ and $\mathcal{D}$, amortized variational scaling learns a mapping approximated by a neural network $G_\beta$, from the task $\mathcal{D}_i$ to the distribution parameters $\{\mu_i,\sigma_i\}$ of $\alpha_i$. 

By leveraging the re-parameterization trick, we obtain the objective function of amortized variational scaling:
\begin{align}\label{finalloss_amor}
\mathcal{L}(\beta,\theta;\mathbf{S})&=-\sum_{i=1}^{n}\sum_{j=m+1}^{m+q}\log p_\theta(y_{i,j}|x_{i,j},\mathcal{D}_i^{tr},g_{\mu_i,\sigma_i}(\epsilon_i))\nonumber\\
&+\log\frac{\sigma_0}{{\sigma_i}^2}+\frac{{\sigma_i}^2+(\mu_i-\mu_0)^2}{2\sigma_0^2},
\end{align}
where $g_{\mu_i,\sigma_i}(\epsilon_i)=\sigma_i\epsilon_i+\mu_i$. Note that the local parameters $\{\mu_i,\sigma_i\}$ are functions of $\beta$, i.e., $\{\mu_i,\sigma_i\}=G_\beta(\mathcal{D}_i)$. We iteratively update $\beta$ and $\theta$ by minimizing the loss function (\ref{finalloss_amor}) during meta-training  

During meta-testing, for each task, the generator produces a variational distribution's parameters and we still use the mean vector as the metric scaling vector for inference.
\subsubsection{ Auxiliary loss.}
To learn the mapping $G_\beta$ from a set $\mathcal{D}_i$ to the variational parameters of the local random variable $\alpha_i$, we compute the mean vector of the embedded queries and the embedded supports as the task prototype to generate the variational parameters. A problem is that the embeddings are not ready to generate good scaling parameters during early epochs.  Existing approaches including co-training~\cite{oreshkin2018tadam} and pre-training~\cite{li2019finding} can alleviate this problem at the expense of computational efficiency. They pre-train or co-train an auxiliary supervised learning classifier in a traditional supervised manner over the meta-sample $\mathbf{S}$, and then apply the pre-trained embeddings to generate the task-specific parameters and fine-tune the embeddings during meta-training. Here, we propose an end-to-end algorithm which can improve training efficiency in comparison with pre-training or co-training. We optimize the following loss function~(\ref{loss_lamb}) where an auxiliary weight $\lambda$ is used instead of minimizing~(\ref{finalloss_amor}) in Algorithm 2 , i.e.,
\begin{align}\label{loss_lamb}
\mathcal{L}_\lambda(\beta,\theta;\mathbf{S})=(1-\lambda)\mathcal{L}(\beta,\theta;\mathbf{S})+\lambda\mathcal{L}(\theta;\mathbf{S}),
\end{align}  
where $\mathcal{L}(\theta;\mathbf{S})=-\sum_{i=1}^{n}\sum_{j=m+1}^{m+q}\log p_\theta(y_{i,j}|x_{i,j},\mathbf{1})$, i.e., no scaling is used. Given a decay step size $\gamma$, $\lambda$ starts from $1$ and linearly decays to $0$ as the number of epochs increases, i.e., $\lambda=\lambda-1/\gamma$. During the first epochs, the weight of the gradients $\frac{\partial\mathcal{L}(\theta;\mathbf{S})}{\partial\theta}$ is high and the algorithm learns the embeddings of PN. As the training proceeds, $\beta$ is updated to tune the learned embedding space. See the details in Algorithm \ref{algorithm2}. 
    \begin{algorithm}[t]
    	\DontPrintSemicolon		
    	\KwInput{Meta-sample $\{\mathcal{D}_i\}_{i=1}^n$, learning rates $l_\theta$, $l_\beta$, prior $\mu_0,\sigma_0$ and step size $l_\lambda$.}
    	\KwInitialize{$\beta$ and $\theta$ randomly, $\lambda=1$.}
    	\For{$i$ in $\{1,2,\dots,n\}$}
    	{$\mathbf{C}_i=\frac{1}{m+q}\sum_{j=1}^{m+q}\phi_\theta(x_{i,j})$
    		
    	\tcp*{Compute the task prototype.}
    	
    	\textcolor{blue}{${\mu_i,\sigma_i}=G_\beta(\mathbf{C}_i)$\\
    		$\epsilon_i\sim\mathcal{N}(\mathbf{0,I}),\alpha_i=\sigma_i\odot\epsilon_i+\mu_i$} 
    	
    	\tcp*{Generate $\mu_i$ and $\sigma_i$ for $i^{th}$ task.}
    	
    		\For{$k$ in $\{1,2,\dots,K\}$}
    		{	
    			$\textbf{c}_k=\frac{1}{N}\sum_{z_{i,j}\in\mathcal{D}_i^{tr},y_{i,j}=k}\phi_\theta(x_{i,j})$ }
    		
    		\For{$j$ in $\{m+1,m+2,\dots,m+q\}$}
    		{\textcolor{blue}{$d(x_{i,j},\mathbf{c}_k)
    				=(\phi_\theta(x_{i,j})-\mathbf{c}_{k})^{T}\bigl(\begin{smallmatrix}
    				\alpha_i^1 &  &  & \\ 
    				&  \alpha_i^2&  & \\ 
    				&  & \dots & \\ 
    				&  &  & \alpha_i^M
    				\end{smallmatrix}\bigr)(\phi_\theta(x_{i,j})-\mathbf{c}_{k}).$}
    		}
    		
    		$\theta=\theta-l_\theta*\nabla_\theta \mathcal{L}_{\lambda}(\beta,\theta;\mathcal{D}_i)$
    		
    		\tcp*{Update the model parameters $\theta$.}
    		\textcolor{blue}{$\beta=\beta-l_\beta*\nabla_\beta \mathcal{L}_{\lambda}(\beta,\theta;\mathcal{D}_i)$}
    		
    		\tcp*{Update the parameters $\beta$ of the generator.}
    	
    		\textcolor{blue}{\If{$\lambda\neq 0$}{$\lambda=\lambda-l_\lambda$}}
    		} 
    	\caption{Dimensional Amortized Variational Scaling for Prototypical Networks}
    	\label{algorithm2}
    \end{algorithm}
\section{5. Experiments} 

To evaluate our methods, we plug them into two popular algorithms, prototypical networks (PN) \cite{snell2017prototypical} and TADAM \cite{oreshkin2018tadam}, implemented by both Conv-4 and ResNet-12 backbone networks. To be elaborated later, Table \ref{key result} shows our main results in comparison to state-of-the-art meta-algorithms, where it can been that our dimensional stochastic variational scaling algorithm outperforms other methods substantially. For TADAM, we incorporate our methods into TADAM in conjunction with all the techniques proposed in their paper and still observe notable improvement.

\begin{table*}[h]
	\centering
	\begin{tabular}{p{1.5cm}|c|c|c}
		
		\specialrule{1pt}{0pt}{0pt}
		\multicolumn{4}{ c }{\textbf{\textit{mini}ImageNet test accuracy}} \\
		\hline
		\textbf{Backbones}&\textbf{Model}&$5$-way $1$-shot & $5$-way $5$-shot\\
		\hline
		\multirow{8}{*}{Conv-4}
		&Matching networks \citep{vinyals2016matching} & $43.56\pm 0.84$& $55.31\pm 0.73$  \\
		&Relation Net \citep{sung2018learning} & $  50.44\pm 0.82$& $ 65.32\pm 0.70$ \\
		&Meta-learner LSTM \citep{ravi2016optimization} & $43.44\pm 0.77$ & $60.60\pm 0.71$\\
		&MAML \citep{finn2017model} & $48.70\pm   1.84$& $ 63.11\pm 0.92$ \\
		&LLAMA \citep{grant2018recasting} & $49.40\pm 1.83$& $-$ \\
		&REPTILE \citep{nichol2018reptile} & $49.97\pm 0.32$& $65.99\pm 0.58$ \\
		&PLATIPUS \citep{finn2018probabilistic} & $50.13\pm  1.86$& $-$ \\

		\hline
		\multirow{8}{*}{ResNet-12}
		&adaResNet \citep{munkhdalai2017rapid} & $56.88\pm 0.62$& $ 71.94\pm  0.57$ \\
		&SNAIL \citep{mishra2017simple} & $55.71\pm 0.99$& $68.88\pm 0.92$ \\
		&TADAM \citep{oreshkin2018tadam} & $58.50\pm0.30$& $ 76.70\pm  0.30$ \\
		&TADAM Euclidean + D-SVS (\textbf{ours}) & $\mathbf{60.16\pm 0.47}$& $\mathbf{77.25\pm 0.15}$  \\
		\cline{2-4}
		&PN Euclidean \citep{snell2017prototypical} * & $53.89\pm 0.38$& $ 73.59\pm  0.48$ \\
		&PN Cosine \citep{snell2017prototypical} *  &$52.31\pm0.83$&$70.74\pm0.24$\\
		&PN Euclidean + D-SVS (\textbf{ours}) *& $\mathbf{55.30\pm 0.08}$& $\mathbf{74.93\pm 0.31}$ \\
		&PN cosine + D-SVS (\textbf{ours}) *& $\mathbf{56.09\pm 0.19}$& $\mathbf{74.46\pm 0.17}$ \\
		\specialrule{1pt}{0pt}{0pt}
	\end{tabular}
	\caption{Test accuracies of 5-way classification tasks on \textit{mini}ImageNet using Conv-4 and ResNet-12 respectively. * indicates results by our re-implementation.}
	\label{key result}
\end{table*}
\begin{table*}[!ht]
	\begin{center}
		\begin{tabular}{ c| c c|c c  } 
			\specialrule{1pt}{0pt}{0pt}
			&\multicolumn{2}{ c| }{5-way 1-shot}&\multicolumn{2}{ c }{5-way 5-shot}\\
			&Euclidean&Cosine&Euclidean& Cosine\\
			\hline
			PN&$44.15\pm 0.39$&$42.20\pm 0.66$&$65.49\pm0.53$&$60.91 \pm 0.50$\\
			\hline
			PN + SVS&$47.84\pm 0.16$&$48.43\pm 0.20$&$66.86\pm 0.06$&$67.02\pm 0.14$\\ 
			PN + D-SVS&$49.01\pm 0.39$&$49.20\pm 0.05$&$67.40\pm 0.32$&$67.33\pm0.23$\\
			PN + D-AVS&$\mathbf{49.10\pm0.14}$&$\mathbf{49.34\pm 0.29}$&$\mathbf{68.04\pm 0.16}$&$\mathbf{67.83\pm 0.16}$ \\
			
			\specialrule{1pt}{0pt}{0pt}
		\end{tabular}
	\end{center}
	\caption{Results of prototypical networks (the first row) and prototypical networks with SVS, D-SVS and D-AVS respectively by our re-implementation using Conv-4.}
	\label{ablation_vi}
\end{table*}
\subsection{5.1. Dataset and Experimental Setup}

\subsubsection{\textit{mini}ImageNet.} The \textit{mini}ImageNet \cite{vinyals2016matching} consists of 100 classes with 600 images per class. We follow the data split suggested by \citet{ravi2016optimization}, where the dataset is separated into a training set with 64 classes, a testing set with 20 classes and a validation set with 16 classes.

\subsubsection{Model architecture.} 

To evaluate our methods with different backbone networks, we re-implement PN with the Conv-4 architecture proposed by \citet{snell2017prototypical} and the ResNet-12 architecture adopted by \citet{oreshkin2018tadam}, respectively. 

The Conv-4 backbone contains four convolutional blocks, where each block is sequentially composed of a 3 $\times$ 3 kernel convolution with 64 filters, a batch normalization layer, a ReLU nonlinear layer and a 2 $\times$ 2 max-pooling layer.

The ResNet-12 architecture contains 4 Res blocks, where each block consists of 3 convolutional blocks followed by a 2 $\times$ 2 max-pooling layer. 

\subsubsection{Training details.}
We follow the episodic training strategy proposed by \citep{vinyals2016matching}. In each episode, $K$ classes and $N$ shots per class are selected from the training set, the validation set or the testing set. For fair comparisons, the number of queries, the sampling strategy of queries, and the testing strategy are designed in line with PN or TADAM.

For Conv-4, we use Adam optimizer with a learning rate of $1e-3$ without weight decay. The total number of training episodes is $20,000$ for Conv-4. 
And for ResNet-12, we use SGD optimizer with momentum $0.9$, weight decay $4e-4$ and $45,000$ episodes in total. 
The learning rate is initialized as $0.1$ and decayed  $90\%$ at episode steps $15000$, $30000$ and $35000$. 
Besides, we use gradient clipping when training ResNet-12. The reported results are the mean accuracies with $95\%$ confidence intervals estimated by $5$ runs.

We normalize the embeddings before computing the distances between them. As shown in Eq.~(\ref{gradient of mu}) and (\ref{gradient of sigma}), the gradient magnitude of variational metric scaling parameters is proportional to the norm of embeddings. 
Therefore, to foster the learning process of these parameters, we adopt a separate learning rate $l_\psi$ for all variational metric scaling parameters.
\subsection{5.2. Evaluation}
The effectiveness of our proposed methods is illustrated in Table \ref{ablation_vi} progressively, including stochastic variational scaling (SVS), dimensional stochastic variational scaling (D-SVS) and dimensional amortized variational scaling (D-AVS). On both $5$-way $5$-shot and $5$-way $1$-shot classification, noticeable improvement can be seen after incorporating SVS into PN. Compared to SVS, D-SVS is more effective, especially for $5$-way $1$-shot classification. D-AVS performs even better than D-SVS by considering task-relevant information.

\subsubsection{Performance of SVS.}
We study the performance of SVS by incorporating it into PN. We consider both $5$-way and $20$-way training scenarios. The prior distribution of the metric scaling parameter is set as $p(\alpha)=\mathcal{N}(1,1)$ and the variational parameters are initialized as $\mu_{init}=100$, $\sigma_{init}=0.2$. The learning rate is set to be $l_\psi=1e-4$. 

Results in Table \ref{single_scaling_exp} show the effect of the metric scaling parameter (SVS). Particularly, significant improvement is observed for the case of PN with cosine similarity and for the case of $5$-way $1$-shot classification. Moreover, it can be seen that with metric scaling there is no clear difference between the performance of Euclidean distance and cosine similarity.

We also compare the performance of a fixed $\sigma=0.2$ with a trainable $\sigma$. We add a shifted ReLU activation function ($x = \max \{1e-2, x\}$) on the learned $\sigma$ to ensure it is positive. Nevertheless, in our experiments, we observe that the training is very stable and the variance is always positive even without the ReLU activation function. We also find that there is no significant difference between the two settings. Hence, we treat $\sigma$ as a fixed hyperparameter in other experiments.

\begin{table*}
	\centering
	\begin{tabular}{ c|c c|c c } 
		\specialrule{1pt}{0pt}{0pt}
		& \multicolumn{2}{ c| }{5-way 1-shot}&\multicolumn{2}{ c }{5-way 5-shot}\\
		&5-way training&20-way training&5-way training&20-way training\\
		\hline
		PN Euclidean &$44.15\pm 0.39$&$48.05\pm 0.47$& $65.49\pm0.53$&$67.32\pm 1.20$\\ 
		PN Cosine &$42.20\pm 0.66$&$46.75\pm0.18$& $60.91\pm0.50$&$66.28\pm 0.14$\\ 
		\hline
		PN Euclidean + SVS  ($\sigma=0.2$) &$47.84\pm 0.16$&$51.15\pm 0.16$& $66.86\pm 0.06$&$68.00\pm 0.22$\\ 
		PN Cosine + SVS ($\sigma=0.2$) &$48.12\pm 0.13$&$51.74\pm0.13$& $66.95\pm0.78$&$67.88\pm 0.10$\\ 
		PN Euclidean + SVS (learned $\sigma$) &$48.28\pm0.14$&$51.36\pm0.15$& $66.84\pm0.30$&$67.80\pm 0.06$\\ 
		PN Cosine + SVS (learned $\sigma$)  &$48.43\pm0.20$&$51.68\pm0.18$& $67.02\pm0.14$&$67.72\pm 0.16$\\ 
		\specialrule{1pt}{0pt}{0pt}
	\end{tabular}
	\caption{Results of prototypical networks and prototypical networks with SVS by our re-implementation using Conv-4.}
	\label{single_scaling_exp}
\end{table*}
\begin{table*}
	\begin{center}
		\begin{tabular}{c c| c c| c c } 
			\specialrule{1pt}{0pt}{0pt}
			&& \multicolumn{2}{ c| }{5-way 1-shot}&\multicolumn{2}{ c }{5-way 5-shot}\\
			Auxiliary training&Prior&Euclidean& Cosine&Euclidean&Cosine\\
			\hline
			&&$47.79\pm 0.10$&$47.45\pm 0.17$&$66.26\pm 0.48$&$66.03\pm0.34$\\
			\hline
			&$\checkmark$&$48.12\pm 0.55$&$47.49\pm 0.26$&$66.69\pm 0.25$&$66.43\pm 0.38$\\
			$\checkmark$&&$48.56\pm0.44$&$49.13\pm0.32$&$67.11\pm0.14$&$67.23\pm 0.19$\\
			\hline
			$\checkmark$&$\checkmark$&$\mathbf{49.10\pm0.14}$&$\mathbf{49.34\pm0.29}$&$\mathbf{68.04\pm 0.16}$&$\mathbf{67.83\pm0.16}$\\
			\specialrule{1pt}{0pt}{0pt}
		\end{tabular}
	\end{center}
	\caption{Ablation study of prototypical networks with D-AVS by our re-implementation using Conv-4.}
	\label{ablation_amortized}
\end{table*}

\subsubsection{Performance of D-SVS.}  
We validate the effectiveness of D-SVS by incorporating it into PN and TADAM, with the results shown in Table \ref{key result} and Table \ref{ablation_vi}. On 5-way-1-shot classification, for PN, we observe about $4.90\%$ and $1.41\%$ absolute increase in test accuracy with Conv-4 and ResNet-12 respectively; for TADAM, $1.66\%$ absolute increase in test accuracy is observed. The learning rate for D-SVS is set to be $l_\psi=16$. Here we use a large learning rate since the gradient magnitude of each dimension of the metric scaling vector is extremely small after normalizing the embeddings.

\subsubsection{Performance of D-AVS.} We evaluate the effectiveness of D-AVS by incorporating it into PN. We use a multi-layer perception (MLP) with one hidden layer as the generator $G_\beta$. The learning rate $l_\beta$ is set to be $1e-3$. In Table \ref{ablation_vi}, on both $5$-way $1$-shot and $5$-way $5$-shot classification, we observe about $1.0\%$ absolute increase in test accuracy for dimensional amortized variational scaling (D-AVS) over SVS with a single scaling parameter. In our experiments, the hyperparameter $\gamma$ is selected from the range of $[100,150]$ with 200 training epochs in total.

\subsubsection{Ablation study of D-AVS.} To assess the effects of the auxiliary training strategy and the prior information, we provide an ablation study as shown in Table \ref{ablation_amortized}. Without the auxiliary training and the prior information, D-AVS degenerates to a task-relevant weight generating approach \cite{lai2018task}. Noticeable performance drops can be observed after removing the two components. Removing either one of them also leads to performance drop, but not as significant as removing both. The empirical results confirm the necessity of the auxiliary training and a proper prior distribution for amortized variational metric scaling.

\subsection{5.3 Robustness Study}
We also design experiments to show: 1) The convergence speed of existing methods does not slow down after incorporating our methods; 2) Given the same prior distribution, the variational parameters converge to the same values in spite of different learning rates and initializations.

\begin{figure}[h]
	\centering
	\subfigure[$5$-way $1$-shot]{
		\begin{minipage}[c]{0.45\columnwidth}
			\includegraphics[width=1\columnwidth]{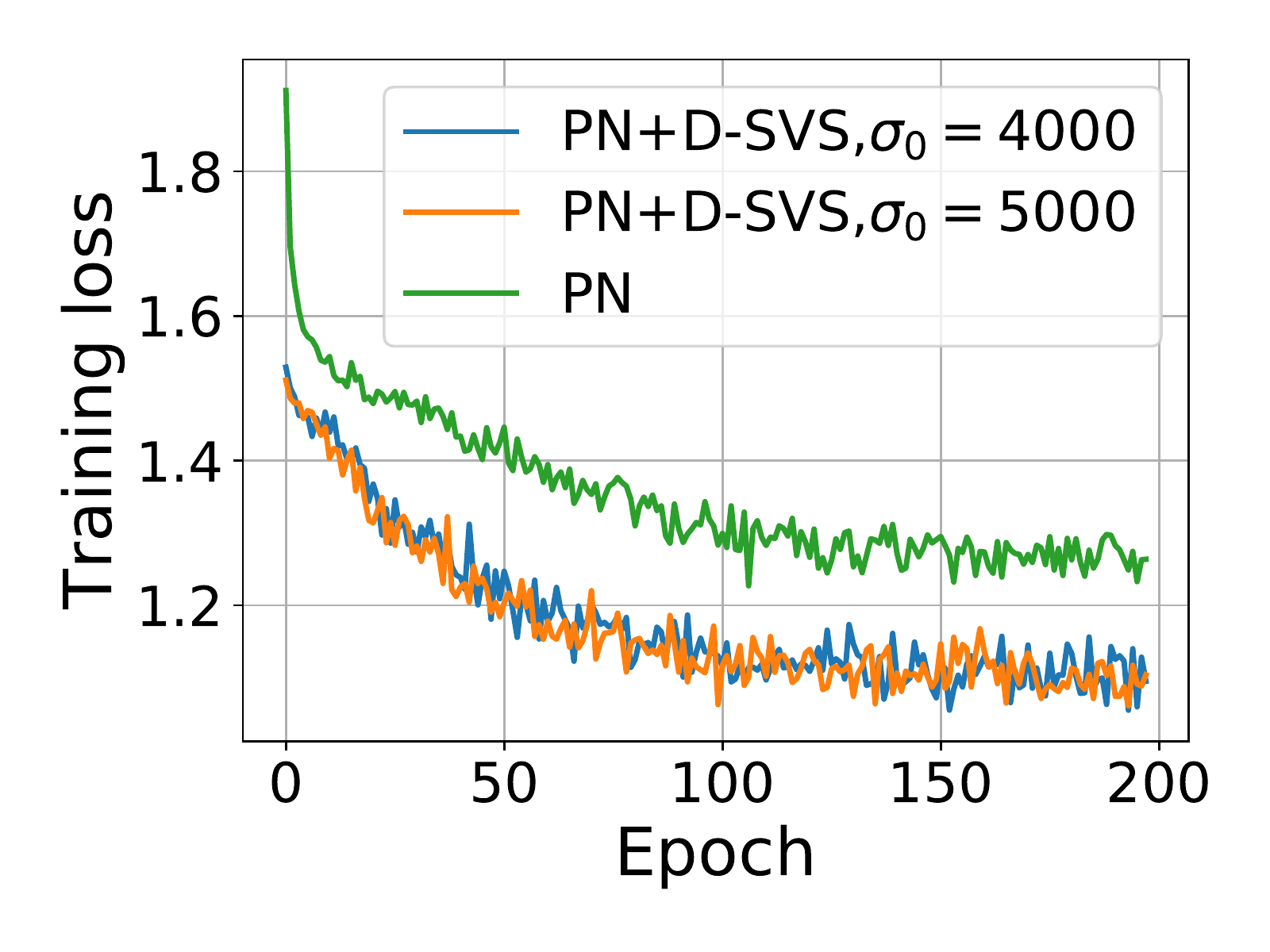}
		\end{minipage}
		\label{2a}
	}
	\subfigure[$5$-way $5$-shot]{
		\begin{minipage}[a]{0.45\columnwidth}
			\includegraphics[width=1\columnwidth]{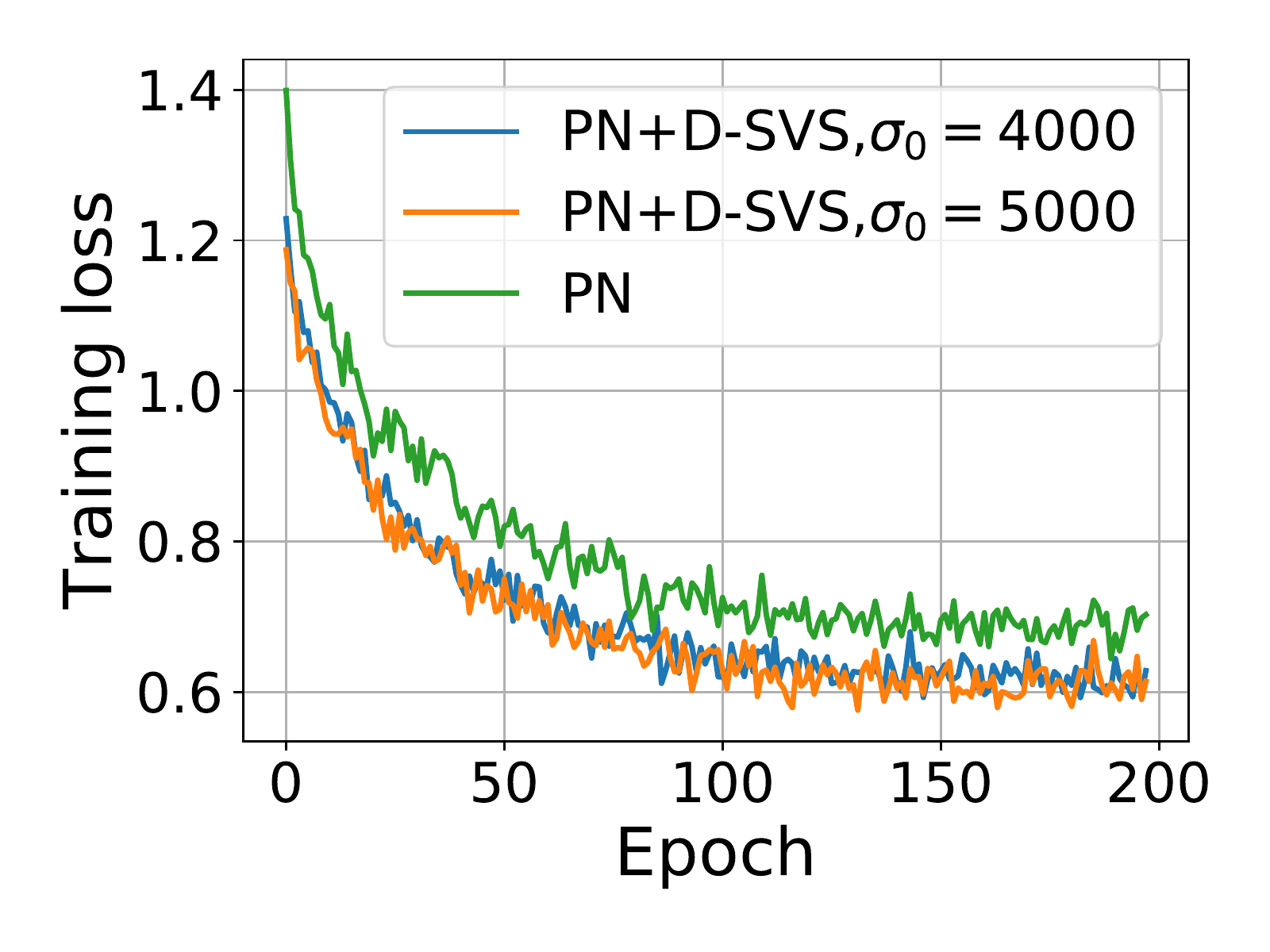}
		\end{minipage}
		\label{2b}
	}
	\caption{Learning curves of prototypical networks and prototypical networks with D-SVS.}
	\label{convergence_rate}
\end{figure}

For the iterative update of the model parameters $\theta$ and the variational parameters $\psi$, a natural question is whether it will slow down the convergence speed of the algorithm. Figure \ref{convergence_rate} shows the learning curves of PN and PN+D-SVS on both 5-way 1-shot and 5-way 5-shot classification. It can be seen that the incorporation of SVS does not reduce the convergence speed.

\begin{figure}[h]
	\centering
	\subfigure[$\mu_0=1,l_\psi=1e-3$]{
		\begin{minipage}[c]{0.45\columnwidth}
			\includegraphics[width=1\columnwidth]{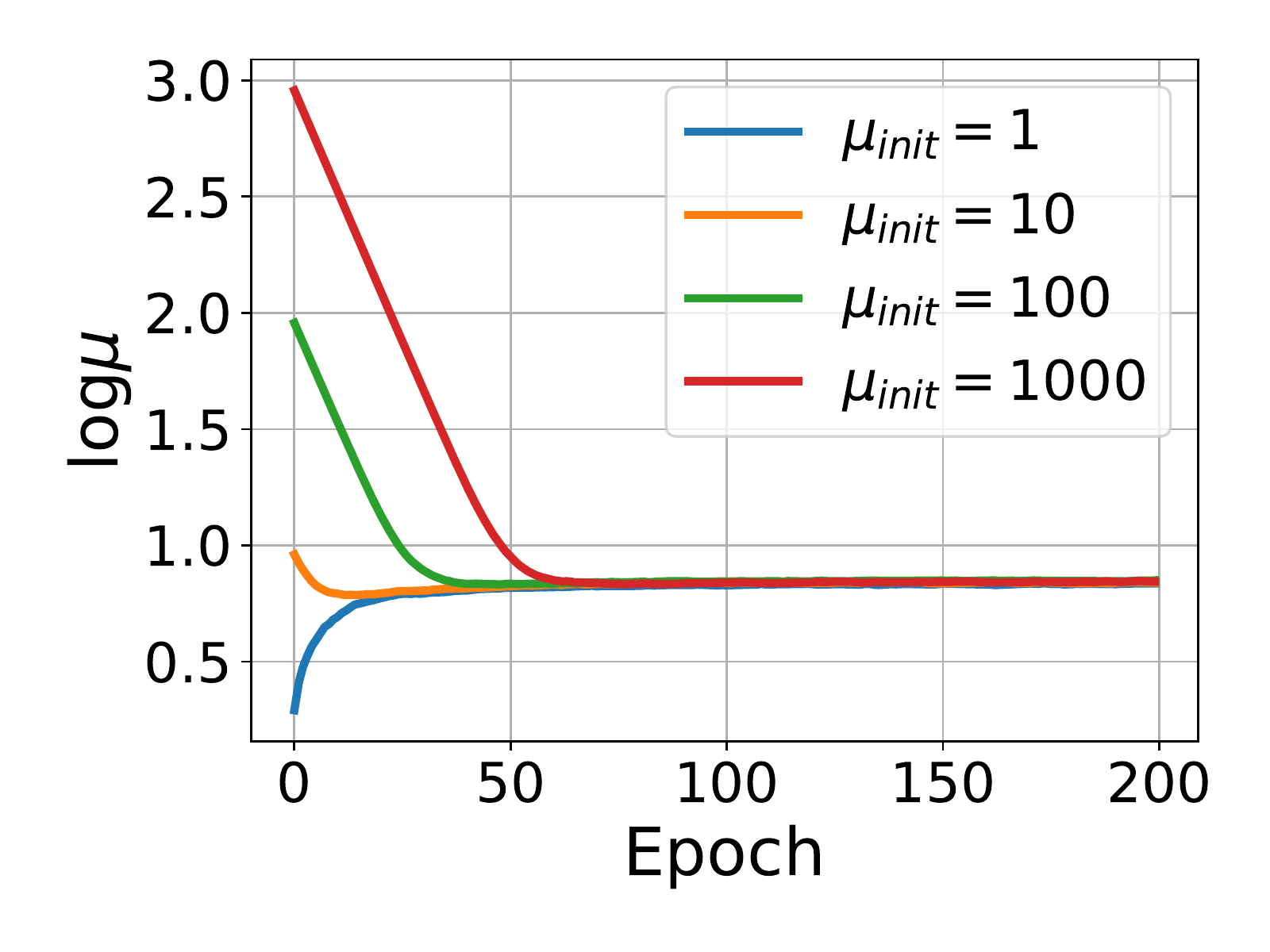}
		\end{minipage}
		\label{mu_a}
	}
	\subfigure[$\mu_0=1,\mu_{init}=100$]{
		\begin{minipage}[a]{0.45\columnwidth}
			\includegraphics[width=1\columnwidth]{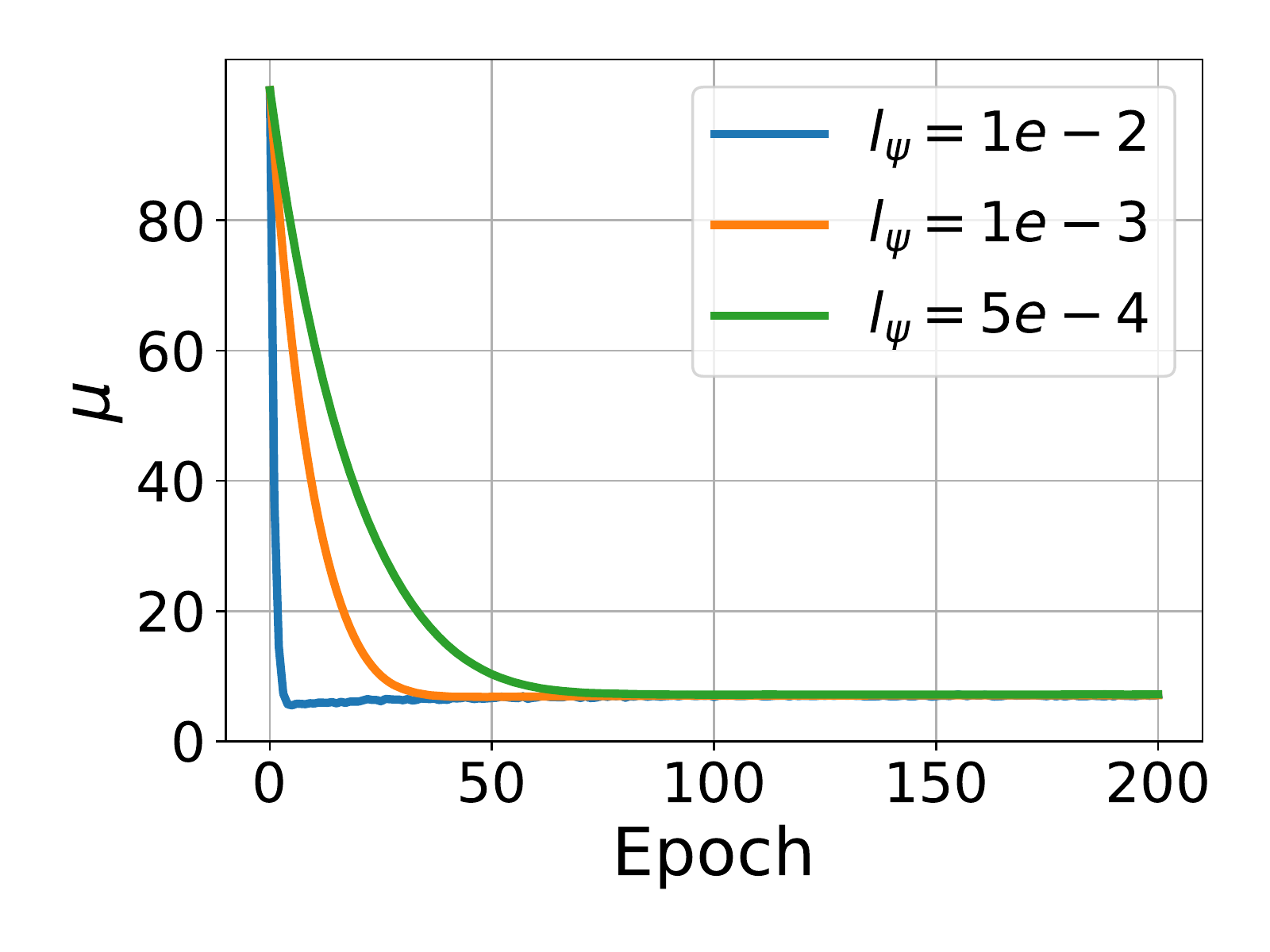}
		\end{minipage}
		\label{mu_b}
	}
	\caption{Learning curves of $\mu$ (a) for different initializations and (b) for different learning rates.}
\end{figure}

We plot the learning curves of the variational parameter $\mu$ w.r.t. different initializations and different learning rates $l_\psi$.
Given the same prior distribution $\mu_0=1$, Fig.~\ref{mu_a} shows that the variational parameter $\mu$ with different initializations will converge to the same value. Fig.~\ref{mu_b} shows that $\mu$ is robust to different learning rates.

\section{Conclusion}
In this paper, we have proposed a generic variational metric scaling framework for metric-based meta-algorithms, under which three efficient end-to-end methods are developed. To learn a better embedding space to fit data distribution, we have considered the influence of metric scaling on the embedding space by taking into account data-dependent and task-dependent information progressively. Our methods are lightweight and can be easily plugged into existing metric-based meta-algorithms to improve their performance. 

\section{Acknowledgements}
We would like to thank the anonymous reviewers for their helpful comments. This research was supported by the grants of DaSAIL projects P0030935 and P0030970 funded by PolyU (UGC).

\bibliographystyle{aaai}
\bibliography{reference}
\clearpage
\appendix

\section{Appendix A. SVS}

\subsection{A.1. Comparison with a Special Case}
\citet{ranjan2017l2} proposed to train the single scaling parameter together with model parameters. Their method can be seen as a special case of our stochastic variational scaling method SVS under the conditions of  $q_{\psi}(\alpha)=\mathcal{N}(\mu,0)$, $\sigma_0\rightarrow\infty$ and $l_\psi=l_\theta$. 
We compare our method with theirs by varying the initialization of $\mu$ ($\mu_{init}$).

Noticeably, our method achieves absolute improvements of $2.86\%$, $1.77\%$, $0.73\%$ and $2.5\%$ for four different initializations respectively. 
As shown in Table \ref{trainable}, our method is stable w.r.t. the initialization of $\mu$, thanks to the prior information introduced in our Bayesian framework which may counteract the influence of initialization.

\begin{table}[!h]
	\begin{center}
		\begin{tabular}{c|c|c|c|c}
			\specialrule{1pt}{0pt}{0pt}	
			\diagbox{Method}{$\mu_{init}$}&1&10&100&1000\\
			\hline
			PN+SVS&$66.45$&$66.95$&$67.02$&$66.72$\\
			PN (Training together)&$63.59$&$65.18$&$66.29$&$64.22$\\
			\specialrule{1pt}{0pt}{0pt}	
		\end{tabular}
	\end{center}
	\caption{Comparison of PN (Training together) and PN+SVS implemented by Conv-4 backbone.}
	\label{trainable}
\end{table}

\subsection{A.2. Sensitivity to the Prior and Initialization} 
In Bayesian framework, the prior distribution has a significant impact on learning posterior distribution. For stochastic variational inference, initialization is another key factor for learning the variational parameters. Here, we conduct experiments of PN+SVS with different prior distributions and initializations. The results of $5$-way $5$-shot classification are summarized in Table \ref{sensitivity}. It can be observed that our method is not sensitive to the prior and initialization as long as either one of them is not too small. 

\section{Appendix B. D-SVS}

\subsection{B.1. Distributions of the Mean Vector $\mu$}
\begin{figure*}[t]
	\centering
	\includegraphics[width=0.8\textwidth]{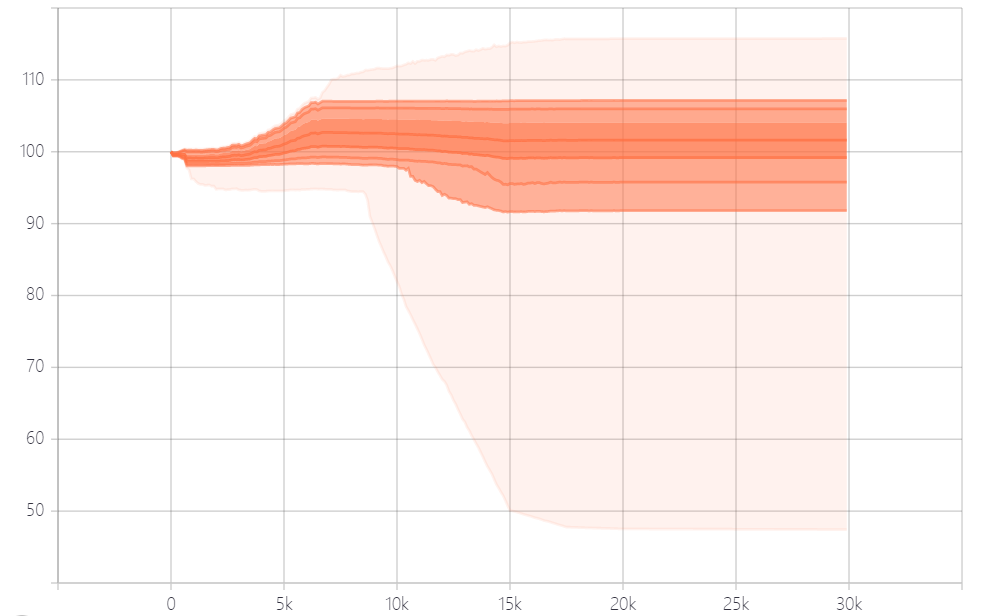}
	\includegraphics[width=0.8\textwidth]{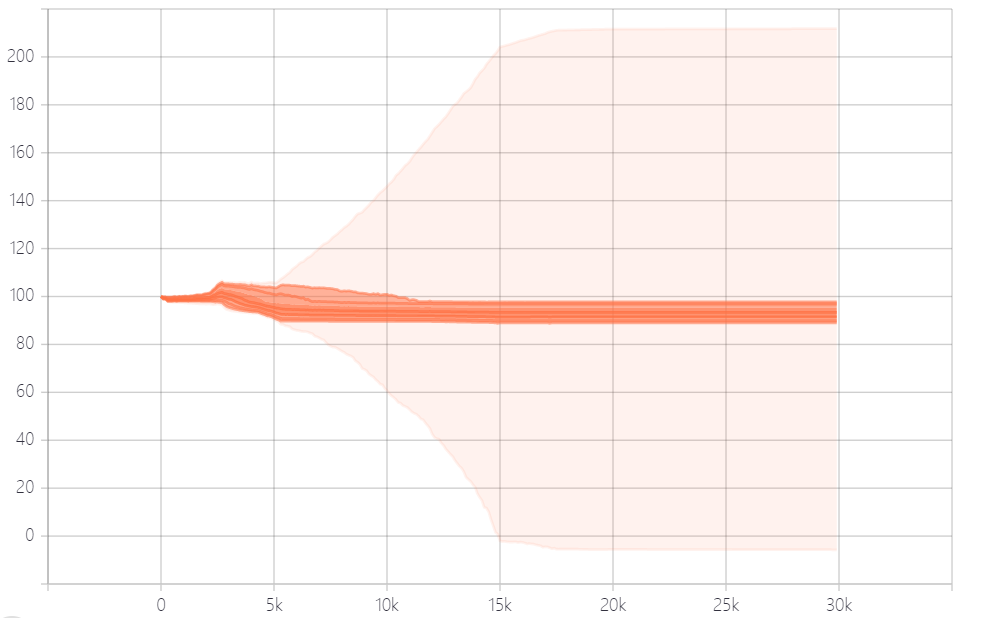}
	\caption{Distributions of the learned $\mu$ w.r.t. the number of training steps. The horizontal and vertical axes are the number of training steps and values of $\mu$, respectively. The top is for $5$-way $1$-shot classification and the bottom is for $5$-way $5$-shot.}
	\label{changing_mu}
\end{figure*}

Figure \ref{changing_mu} illustrates the distributions of the mean vector $ {\mu}=(\mu^1,\mu^2,\dots,\mu^M)$ during the meta-training procedure of $5$-way $1$-shot and $5$-way $5$-shot classification respectively. Darker colour means more frequent occurrence.

At step $0$, all dimensions of $\mu$ are initialized as $100$. They diverge as the meta-training proceeds, which shows D-SVS successfully learns different scaling parameters for different dimensions. It is also worth noting that for both tasks, the distribution of $\mu$ converges eventually (after $15k$ steps).

\section{Appendix C. D-AVS}
\subsection{C.1. Viewing the Learned Metric Scaling Parameters of PN+D-AVS.} 
D-AVS generates variational distributions for different tasks, from which the task-specific scaling parameters are sampled. Below we print out the  scaling vectors learned by D-AVS on two different testing tasks for 5-way 5-shot classification, where only the first ten dimensions are displayed. It can be seen that D-AVS successfully learns tailored scaling vectors for different tasks.

Scaling parameters for Task 1: [64.9170, 22.4030, 13.4468, 2.3949, 28.2470, 29.7770, 54.4221, 60.9279, 2.3008, 147.5304].

Scaling parameters for Task 2: [60.2564, 21.1672, 12.8457, 2.3603, 26.6194, 27.9963, 50.6127, 56.5965, 2.2672, 135.2510].

\section{Appendix D. Implementation Details}
\subsection{D.1. Sampling from the Variational Distribution}
We adopt the following sampling strategy for the proposed three approaches. For meta-training, we sample once per task from the variational distribution for the metric scaling parameter; for meta-testing, we use the mean of the learned Gaussian distribution as the metric scaling parameter. The computational overhead is very small and can be ignored.
\begin{table*}[t]
	\centering
	\begin{tabular}{c|cccc}
		\specialrule{1pt}{0pt}{0pt}
		\diagbox{$\mu_0$}{$\mu_{init}$}&$1$&$10$&$100$&$1000$\\
		\hline
		$1$&$60.25\pm0.70$&$63.00\pm0.34$&$66.86\pm0.06$&$66.26\pm0.24$\\
		$10$&$63.95\pm0.24$&$65.89\pm0.32$&$66.79\pm0.55$&$66.34\pm0.22$\\
		$100$&$65.94\pm0.42$&$66.95\pm0.27$&$67.02\pm0.38$&$66.43\pm0.10$\\
		$1000$&$66.45\pm0.17$&$66.66\pm0.23$&$66.88\pm0.16$&$66.72\pm0.28$\\
		\specialrule{1pt}{0pt}{0pt}
	\end{tabular}
	\caption{Results of PN+SVS w.r.t. different initializations and priors implemented by Conv-4.}
	\label{sensitivity}
\end{table*}

\end{document}